\documentclass[lettersize,journal]{IEEEtran}
\usepackage{amsmath,amsfonts}
\usepackage{algorithmic}
\usepackage{algorithm}
\usepackage{array}
\usepackage[caption=false,font=normalsize,labelfont=sf,textfont=sf]{subfig}
\usepackage{textcomp}
\usepackage{stfloats}
\usepackage{url}
\usepackage{verbatim}
\usepackage{graphicx}
\usepackage{cite}
\usepackage{hyperref}
\usepackage{multirow}
\usepackage{calligra}
\usepackage{algorithm}
\usepackage{amsmath}
 \usepackage{algorithmic}

\begin{document}

\title{Attribute reduction algorithm of rough sets based on spatial optimization}

\author{Xu-chang Guo,~Hou-biao Li} 



\maketitle

\begin{abstract}
Rough set is one of the important methods for rule acquisition and attribute reduction. The current goal of rough set attribute reduction focuses more on minimizing the number of reduced attributes, but ignores the spatial similarity between reduced and decision attributes, which may lead to problems such as increased number of rules and limited generality. In this paper, a rough set attribute reduction algorithm based on spatial optimization is proposed. By introducing the concept of spatial similarity, to find the reduction with the highest spatial similarity, so that the spatial similarity between reduction and decision attributes is higher, and more concise and widespread rules are obtained. In addition, a comparative experiment with the traditional rough set attribute reduction algorithms is designed to prove the effectiveness of the rough set attribute reduction algorithm based on spatial optimization, which has made significant improvements on many datasets. {\textit{Code Availability}}: \url{https://github.com/nkjmnkjm/SRS}.
\end{abstract}

\begin{IEEEkeywords}
Rough set, Spatial similarity, attribute reduction, heuristic algorithm
\end{IEEEkeywords}

\section{Introduction}
\IEEEPARstart{R}{ough} set theory is an important method for learning, simplifying, and revising rules. Since its introduction by Pawlak.Z in 1991 \cite{Zalewski_1996}, rough set theory has become a mathematical tool and method for describing uncertain and incomplete knowledge and data, learning, and induction. It can effectively analyze and process various types of information such as inaccuracy, incompleteness, and inconsistency, and reveal hidden knowledge and potential laws from them.

During the experimental process, traditional methods of attribute reduction in rough set theory tend to select attributes with a large number of attribute values because such data often has stronger classification capabilities, enabling finer divisions of elements in the universe of discourse. However, in some experiments, attributes with many attribute values tend to be more difficult to label summary attributes. For example, in the Zoo animal dataset experiment, the attribute with the most attribute values is the name of the animal, which can directly divide the elements into the universe of discourse into 101 categories, while the decision attribute only has 7 categories. This can directly lead to the generation of 101 rules in the knowledge acquisition process without manually removing this attribute, and the rules have poor generality and cannot reflect the differences in the distinction of decision attribute classes based on conditional attributes.

In this paper, starting from the spatial similarity between decision attribute partitions and conditional attribute partitions, the concept of spatial optimal measurement is defined. This helps to select attributes that better reflect the differences in the distinction of decision attribute classes based on conditional attributes and generate corresponding rules. The main contributions are as follows:

\begin{itemize}
\item{Firstly, cosine similarity is introduced into rough sets as a measure of spatial similarity, and based on this, the concept of spatial optimal measurement is defined.}
\item{Secondly, a spatial optimal rough set attribute reduction algorithm is designed, and the operation process and feasibility of the algorithm are explained.}
\item{Finally, a computational example and a set of comparative experiments are designed to verify the effectiveness of the spatial optimal rough set attribute reduction algorithm.}
\end{itemize}

\section{Related Work}
Since the original Pawlak.Z rough set is based on the indistinguishable binary relation of complete information systems, its application scenarios in real life are relatively strict. Scholars have proposed many new concepts of rough sets to solve practical problems, such as Kryszkiewicz's rough set method for reasoning in incomplete information systems \cite{Kryszkiewicz1998,Kryszkiewicz1999}, hybrid multi-selector method \cite{CHAUDHURI2021}, fuzzy rough set model based on Gaussian kernel approximation by combining Gaussian kernel with fuzzy rough set \cite{HU2010}, and Greco's rough approximation through dominance relations \cite{Greco2001}.

Attribute reduction is one of the important steps in the process of knowledge discovery. It maintains the classification ability of the knowledge base under unchanged conditions and deletes irrelevant or unimportant attributes. The attribute reduction algorithm based on the classic rough set does not require prior knowledge, and can discover the laws of the problem and reduce the attributes only by using the information provided by the data itself. However, it is based on strict equivalence relations and has great limitations, such as poor fault tolerance; it cannot handle incomplete or incompatible decision information tables; and it cannot directly handle numerical attributes, ordered attributes, or dynamically changing attributes in decision information tables. Therefore, many scholars have studied and expanded the rough set theory, successively proposing probability rough sets \cite{pawlak1988rough}, variable precision rough sets \cite{ziarko1993variable}, decision rough sets \cite{yao1992decision}, multi-granularity rough sets \cite{qian2010mgrs}, etc., and exploring incomplete decision information tables \cite{stefanowski1999extension}, incompatible decision information tables \cite{ye2002new}, continuous attribute decision information tables \cite{hu2008efficient}, attribute dynamic change decision information tables \cite{shu2015incremental}, ordered attribute decision information tables \cite{greco2001rough}, etc., achieving good research results. 

In addition, research on attribute reduction algorithms has been carried out based on the extended models of rough sets. The discernibility matrix proposed by Polish mathematician Skowron.A in 1992 \cite{SKOWRON1992} is an important method for dealing with rough set attribute reduction. However, since the minimum reduction problem of information systems has been proven to be an NP-hard problem, scientists have mostly attempted to solve such problems through heuristic algorithms, such as Hu's attribute-oriented rough set approach \cite{Hu1995} in 1995 and Miao's attribute reduction algorithm using mutual information, called MIBARK \cite{Miao1997}. To further improve the speed of the algorithm, scientists have also proposed many accelerated algorithms, such as the feature selection mechanism based on ant colony optimization proposed by Jenshen and Shen to solve the problem of finding the optimal feature subset in fuzzy rough data reduction \cite{JENSEN2005}.

In recent years, researchers have also conducted further studies on the application of rough sets. To address the issue of neighborhood rough set dependency measure functions ignoring higher approximation values, Changzhong Wang et al. proposed the relative neighborhood self-information method \cite{wang2019feature}, as well as Attribute reduction with fuzzy rough self-information measures \cite{wang2021attribute}. Additionally, to address the sensitivity of classical rough set theory to noise, Changzhong Wang et al. also introduced the concept of variable parameters and constructed the Fuzzy Rough Computation Algorithm (FRC) \cite{wang2019fuzzy} for attribute reduction. Of course, in the current era of big data, more attribute reduction methods have been proposed, such as GBNRS \cite{xia2020gbnrs} and FDNRS \cite{sang2021incremental}.

Currently, research on knowledge acquisition and decision support based on rough set theory has been widely used in many fields such as pattern recognition, artificial intelligence, and has become an important method in such fields. For instance, the sine-cosine algorithm based on the information gain fuzzy-rough set \cite{ewees2020improved} is applied to predict osteoporosis.

\section{Preliminary}
This section is used to introduce the basic knowledge about rough sets theory and attribute reduction algorithms.

\subsection{Rough set theory}
Rough set refers to dividing a set into several equivalent classes, where each equivalent class is a rough set.

Let $U\neq\emptyset$ be a finite set composed of all the research objects, called the universe. Any subset $X\subseteq U$ is called a concept or category in $U$. Let $\textbf{R}$ be a set of equivalence relation on $U$, and for any $R\subseteq \textbf{R}$, $AS = \langle U, R \rangle$ is called an approximation space.

Let $U$ be a universe, $R\subseteq \textbf{R}$, for any $X\subseteq U$, the lower approximation ($\underline{R}(X)$) and upper approximation ($\overline{R}(X)$) of $X$ based on equivalence relation $R$ are defined as follows [24]:
\begin{equation}
    \underline{R}(X) = \{x\subseteq U:[x]_{R}\subseteq X\},
\end{equation}
\begin{equation}
    \overline{R}(X) = \{x\subseteq U:[x]_{R}\cap X \neq\emptyset\}.
\end{equation}
$[x]_{R}=\{y\subseteq U:(x,y)\subseteq R\}$ represents the equivalence class of element $X$ defined by equivalence relation $R$ on universe $U$.

Based on the definition of upper and lower approximations, we have obtained the positive region, negative region and boundary region of $X$ as follows:
\begin{equation}
    POS_{R}(X)=\underline{R}(X),
\end{equation}
\begin{equation}
    NEG_{R}(X)=U - \overline{R}(X),
\end{equation}
\begin{equation}
    BND_{R}(X)=\overline{R}(X)-\underline{R}(X).
\end{equation}

Rough set is a commonly used method for handling rule reduction and can also be used to handle rule correction and generation tasks. When we divide the target concept into combinations of sub-concepts, the data constitutes a decision information system $IS = \langle U, AT, V, f \rangle$, where $U$ is the set of elements in the data called the universe, $AT$ is the set of sub-concepts and target concepts, $\forall a\subseteq AT$, $V_a$ is the value domains of $a$, $V$ is the set of value domains of all concepts, and $f:U\times A\rightarrow V$ is an information function, $\forall x\subseteq U, a\subseteq AT$, define $f(x,a)$ to represent the value of $x$ on $a$, then $f(x,a)\subseteq V_a$.

\subsection{Attribute reduction algorithms}
Next, we will introduce three classic rough set attribute reduction algorithms.
\subsubsection{Discernibility matrix}
The discernibility matrix, proposed by Polish mathematician $Skowron.A$ in 1992 \cite{SKOWRON1992}, is an important method for dealing with attribute reduction in rough set theory. Given a compatible decision information system $DIS=\langle U, A=C\cup D, V, f\rangle$, the $i$-th row and $j$-th column of its discernibility matrix $M_D$ is defined as:
\begin{equation}
    \resizebox{.9\linewidth}{!}{
    $M_D(i, j)= \begin{cases}\left\{a_k \mid a_k \in C \wedge a_k\left(x_i\right) \neq a_k\left(x_j\right)\right.\}& ,d\left(x_i\right) \neq d\left(x_j\right) \\ 0 & , d\left(x_i\right)=d\left(x_j\right)\end{cases}$}
\end{equation}
Where $i,j=1,2,…,n$, $a_{i}(x_{j})$ represents the value of element $x_j$ on attribute $a_i$.

From the definition of discernibility matrix, it can be seen that it is a symmetric matrix that only represents the unordered relationship between two elements. When the decision attributes of two elements are the same, its value is zero, indicating that the two elements are consistent from the perspective of decision-making. When the decision attributes of the two elements are different, its value is the set of conditional attributes where the attribute values of the two elements are different. Based on the definition of discernible attributes and the definition of the core in rough set attribute reduction, it can be seen that the union of the sets with a value of 1 in the discernibility matrix is the core of the conditional attribute reduction under the given decision attribute.

\begin{algorithm}[htb]
\renewcommand{\algorithmicrequire}{\textbf{Input:}}
\renewcommand{\algorithmicensure}{\textbf{Output:}}
\caption{ Discernibility matrix\cite{SKOWRON1992}}
\label{alg:Framwork}
\begin{algorithmic}[1] 
\REQUIRE ~~\\ 
   decision system $DIS=\langle U, AT=C \cup D, V,f\rangle$
\ENSURE ~~\\ 
   attribute set $L$
    \STATE $M_{D} \gets DIS$
    \STATE $While \hspace{0.1cm}M_{i j}\hspace{0.2cm}in\hspace{0.2cm} M_{D}:$
    \STATE $\hspace{0.5cm} L_{i j}=\underset{a_i \in M_{i j}}{\vee} a_i$
        \STATE $ end$
    \STATE $L=\wedge_{M_{i j} \neq 0} L_{i j}$
    \STATE $L^{\prime}=\underset{i}{\vee} L_i$
\STATE \textbf{return} $L$
\end{algorithmic}
\end{algorithm}

Algorithm 1 demonstrates the basic flow of the discernibility matrix-based algorithm, which utilizes the discernibility matrix and logical operations to ultimately obtain the result of attribute reduction. Essentially, it transforms the search form of attribute reduction into a logical operation form starting from the attribute core. Generally speaking, due to its clear and simple computation, the discernibility matrix-based algorithm is often used for kernel extraction in rough set attribute reduction. The time complexity of the algorithm is $O((|A|+log|U|)\cdot |U|^2)$.

\subsubsection{An attribute-oriented rough set approach}
An attribute-oriented rough set approach\cite{Hu1995}, was proposed by Hu in 1995. The core step is a heuristic algorithm that starts from the core and adds attributes with the highest dependency degree each time. Given a compatible decision information system $DIS=\langle U, AT=C\cup D, V, f\rangle$, the dependency degree $K(R,D)$ of the conditional attribute set $R$ relative to the decision attribute set $D$ is defined as follows:
\begin{equation}
    K(R,D)=\frac{|POS_{IND(R)}(IND(D))|}{|U|}.
\end{equation}

Attribute dependency reflects the degree of correlation between the conditional attribute set $R$ and the decision attribute set $D$. Based on this, Hu defined the concept of attribute significance $Sig(a,R,D)$, whose basic definition is as follows:
\begin{equation}
    Sig(a,R,D)=\gamma_{R\cup\{a\}}(D)-\gamma_{R}(D).
\end{equation}

Based on the definition of attribute importance, it can be seen that when the attribute importance is higher, it indicates that the newly added attribute $a$ has a greater improvement in the dependency degree $K(R,D)$ of the conditional attribute set $R$ relative to the decision attribute set $D$, and is therefore more important.

\begin{algorithm}[htb]
\renewcommand{\algorithmicrequire}{\textbf{Input:}}
\renewcommand{\algorithmicensure}{\textbf{Output:}}
\caption{ Attribute-oriented rough set approach\cite{Hu1995}}
\label{alg:Framwork}
\begin{algorithmic}[1] 
\REQUIRE ~~\\ 
   decision system $DIS=\langle U, AT=C \cup D, V,f\rangle$
\ENSURE ~~\\ 
   attribute set $R$
    \STATE $R= gets minReduction(D)$
    \STATE $While \hspace{0.1cm}POS_{R}(D)\neq POS_{C}(D):$
    \STATE $\hspace{0.5cm} a \gets max Sig(a,R,D) (a \subseteq C)$
    \STATE $\hspace{0.5cm} R \gets R \cup \{a\}$
        \STATE $ end$
    \STATE $L=\wedge_{M_{i j} \neq 0} L_{i j}$
    \STATE $L^{\prime}=\underset{i}{\vee} L_i$
\STATE \textbf{return} $R$
\end{algorithmic}
\end{algorithm}

Algorithm 2 demonstrates the process of the heuristic algorithm based on attribute importance. It first derives the core of the attribute set, which is typically done using the method of discernibility matrix. If the positive region based on the attribute core is consistent with the positive region based on the conditional attribute set, the core is directly derived as the result of attribute reduction; otherwise, it calculates the attribute with the highest importance in the conditional attribute set, adds it to the core, and continues the calculation until the positive region based on the reduced attribute set is consistent with the positive region based on the conditional attribute set. The reduced attribute set is then derived. The time complexity of the algorithm is composed of two parts: the core computation and the attribute reduction. The time complexity of the core computation is $O(|A|\cdot|U|^2)$, and the time complexity of the reduction part is also $O(|A|\cdot|U|^2)$.

\subsubsection{MIBARK algorithm}
Information entropy is one of the important concepts in information theory, which represents the uncertainty of information. Based on their understanding of the basic theory of rough sets, Miao Duoqian, Wang Jue, and others introduced information entropy into the attribute reduction of rough sets and proposed an attribute reduction algorithm using mutual information, known as the MIBARK algorithm \cite{Miao1997}.

For an attribute set $P$, its information entropy $H(P)$ is defined as:
\begin{equation}
H(P)=-\sum_{i=1}^n p\left(X_i\right) \log p\left(X_i\right).
\end{equation}
Generally, the base of logarithmic functions is 2.

Having the basic definition of information entropy, the information entropy of a decision attribute set partition $Q\left(U / I N D(D)=\left\{Y_1, \ldots, Y_m\right\}\right)$ relative to a conditional attribute set $C\left(U / I N D(C)=\left\{X_1, \ldots, X_m\right\}\right)$ is defined as:
\begin{equation}
H(Q \mid P)=-\sum_{i=1}^n p\left(X_i\right) \sum_{j=1}^m p\left(Y_j \mid X_i\right) \log p\left(Y_j \mid X_i\right).
\end{equation}

After introducing information entropy, the definition of attribute reduction based on information entropy for rough sets can be obtained. In a compatible decision information system $DIS=\langle U, AT=C\cup D, V, f\rangle$, where $P \in C$ and $Q \in P$, the necessary and sufficient condition for $Q$ to be a reduction of $P$ is as follows:
\begin{itemize}
    \item $H(D/Q)=H(D/P)$,
    \item $\vee q \in Q, H(D/Q)\neq H(D/Q-{q})$.
\end{itemize}

After defining the attribute reduction of rough sets based on information entropy, similar to heuristic algorithms based on attribute importance, the concept of attribute importance $SGF(a,R,D)$ can also be defined. Its basic definition is as follows:
\begin{equation}
    SGF(a,R,D)=H(D|R)-H(D|R\cup{a}).
\end{equation}

The greater the importance of the same attribute $SGF(a,R,D)$, the more important the attribute $a$ is based on the conditional attribute set $R$. In the special case when $R=\emptyset$, the importance of the attribute $SGF(a,R,D)$ degenerates into the mutual information between attribute $a$ and the decision attribute set $D$.

\begin{algorithm}[htb]
\renewcommand{\algorithmicrequire}{\textbf{Input:}}
\renewcommand{\algorithmicensure}{\textbf{Output:}}
\caption{MIBARK algorithm\cite{Miao1997}}
\label{alg:Framwork}
\begin{algorithmic}[1] 
\REQUIRE ~~\\ 
   decision system $DIS=\langle U, AT=C \cup D, V,f\rangle$
\ENSURE ~~\\ 
   attribute set $R$
    \STATE $R= gets minReduction(D)$
    \STATE $While \hspace{0.1cm}I(R,D)\neq I(C,D):$
    \STATE $\hspace{0.5cm} a \gets max SGF(a,R,D) (a \subseteq C)$
    \STATE $\hspace{0.5cm} R \gets R \cup \{a\}$
    \STATE $ end$
    \STATE $L=\wedge_{M_{i j} \neq 0} L_{i j}$
    \STATE $L^{\prime}=\underset{i}{\vee} L_i$
\STATE \textbf{return} $R$
\end{algorithmic}
\end{algorithm}

Algorithm 3 demonstrates the process of the MIBARK algorithm. Similar to heuristic algorithms based on attribute importance, it first derives the core of the attribute set, which is typically done using the discernibility matrix approach. When the mutual information based on the attribute core is consistent with the mutual information based on the conditional attribute set, the core is directly derived as the result of attribute reduction. Otherwise, it calculates the attribute with the highest importance in the conditional attribute set, adds it to the core, and continues the calculation until the mutual information based on the reduced attribute set is consistent with the mutual information based on the conditional attribute set. Finally, the reduced attribute set is derived. The time complexity of the algorithm is divided into two parts: the core-seeking part and the attribute reduction part. The time complexity of the core-seeking part is $O(|A|\cdot|U|^2)$, and the time complexity of the attribute reduction part is $O(|A|^2)$.

\section{A spatial optimal rough set attribute reduction algorithm}
Analyzing the three classical rough set attribute reduction algorithms mentioned above, it can be found that the method based on discernibility matrix has a high time complexity, which is not suitable for handling large datasets. Both heuristic algorithms tend to favor attributes with more attribute values, which may result in a larger number of element sets in the partitioning of the reduced set. In subsequent rule learning, the learned rules may be more cumbersome and have poor generalization performance. For example, in the Zoo dataset, if all attributes are directly used for attribute reduction, the resulting reduction is simply the names of animals, resulting in 101 element sets in the partitioning. The rules derived from this reduction are 101 rules, while the classification of the decision attribute set itself only has 7 categories. Obviously, the rules cannot reflect the unified rules between the attribute sets regarding the conditional attributes.

Generally speaking, the goal of rough set attribute reduction is to make the reduced attribute set smaller, but this goal cannot reflect the relationship between conditional attributes and decision attributes. To reflect the relationship between the reduced attribute set and the decision attribute set, spatial similarity is a good discriminant indicator. Obviously, the higher the spatial similarity between the reduced set and the decision attribute set, the closer the generated rules are to the minimum number of rules, and the higher the generalization of the rules. To obtain more concise and generalizable rules, this paper proposes a spatial optimal rough set attribute reduction algorithm (SRS) that considers the spatial structure after partitioning more, aiming to obtain an attribute reduction that is closer to the partitioning of the decision attribute set.

\subsection{Spatial similarity}
In an information system $IS=\langle U, A, V,f\rangle$, the similarity of attribute subsets $B,C \in A$ can be characterized by information entropy or positive region on one hand, and can also be represented by spatial similarity of partitions on the other hand.

\newtheorem{Definition}{\bf Definition}
\begin{Definition}\label{thm1}
A partition $Q\left(U/IND(D)=\left\{Y_1, \ldots, Y_m\right\}\right)$ based on the decision attribute set $D$, and a partition $P\left(U/IND(C)=\left\{X_1, \ldots, X_m\right\}\right)$ based on the conditional attribute $C$, the spatial similarity between the two is measured by the cosine similarity $cos(P,Q)$ after sorting the sets of elements in the partitions in descending order of their cardinality, i.e.,
\begin{equation}
    cos(P,Q)=\frac{P\cdot Q}{\parallel P \parallel \parallel Q \parallel}.
\end{equation}
\end{Definition}

From the above definition, it can be seen that after sorting the cardinalities of the element sets in the partitions from largest to smallest, the two partitions can be regarded as two ordered vectors. The spatial similarity between the two partitions can then be characterized by the cosine similarity. Obviously, due to the definition of partitions as ordered sets from largest to smallest, the spatial similarity $cos(P,Q)$ takes values in the range $(0,1]$. When $cos(P,Q)=1$, it indicates that the two vectors point in the same direction, indicating the highest spatial similarity. The smaller the value of $cos(P,Q)$, the greater the deviation in the directions pointed by the two vectors, and the lower the spatial similarity.

\subsection{Optimal spatial metric}
However, spatial similarity only focuses on the similarity of the ordered set directions of two partitions without considering the similarity of the element sets within the partitions. Therefore, it cannot be directly used for spatial optimization measurement. To address this, the similarity of the internal element sets of the partitions needs to be introduced to assist in defining spatial optimization.

\begin{Definition}
For a partition based on the decision attribute set $D$, denoted as $Q\left(U/IND(D)=\left\{Y_1, \ldots, Y_m\right\}\right)$, and a partition based on the conditional attribute $C$, denoted as $P\left(U/IND(C)=\left\{X_1, \ldots, X_m\right\}\right)$, the spatial similarity between the two partitions is measured by the cosine similarity $cos(P,Q)$ after sorting the cardinalities of the element sets in the partitions from largest to smallest. The positive region of partition $Q$ relative to partition $P$ is denoted as $POS_{P}(Q)$. The spatial optimization measurement between the two partitions, denoted as $SPS(P,Q)$, is defined as follows:
\begin{equation}
    SPS(P,Q)=\alpha cos(C,D)+\beta \frac{ |POS_{P}(Q)|}{|U|}.
\end{equation}
\end{Definition}
In this context, $\alpha$ and $\beta$ are defined parameters, and it is given that $\alpha+\beta=1$.

Having defined spatial optimization, we can analogously define the concept of spatial optimal importance of an attribute, denoted as$SigSPS(a,R,D)$, in alignment with heuristic algorithms based on attribute importance. The basic definition is as follows:
\begin{equation}
    SigSPS(a,R,Q)=SPS(R\cup{a},D)-SPS(R,D).
\end{equation}

The greater the spatial optimal attribute importance $SigSPS(a,R,D)$ is, the more important the attribute $a$ is based on the conditional attribute set $R$.

In the task of attribute reduction in rough sets, it is required that the positive region of the partition $Q$ based on the decision attribute set $D$ should contain all elements within the universe of discourse, i.e., $|POS_P(D)|=|U|$. The spatial optimal attribute importance $SPS(P,Q)$ cannot directly reflect this condition. Therefore, during application, when none of the candidate attributes satisfy $SigSPS(a,R,D)>0$, the element with the largest $\frac{|POS_P(D)|}{|U|}$ is selected.
\subsection{Feasibility analysis}
\newtheorem{property}{\bf Property}
\begin{property}\label{thm3}
For a compatible decision information system $DIS=\langle U, AT=C\cup D, V,f\rangle$, based on the partitioning of the decision attribute set as $U_{D}=U/IND(D)$, for an attribute set $R$ on the conditional attribute set $C$, its partitioning is $U_{R}=U/IND(R)$, and for another attribute set $P=R\cup \{X\}$, its partitioning is $U_{P}=U/IND(P)$, it holds that $|U_{R}|\leq |U_{P}|$.
\end{property}

\begin{IEEEproof}
Obviously, $|U_{R}|\leq |POS_{P}(U_{R})| = |U_{P}|$.
\end{IEEEproof}

Property 1 reflects that when additional attributes are added to the conditional attribute set, the number of elements in the resulting partitioning can only increase, meaning that the cardinality (or the size) of the partition can only increase. Similarly, when conditional attributes are added to the reduction set, the number of derived rules can only increase.

\begin{property}\label{thm1}
For a compatible decision information system $DIS=\langle U, AT=C\cup D, V,f\rangle$, based on the partitioning of the decision attribute set as $U_{D}=U/IND(D)$, for a reduction $R$ on the conditional attribute set $C$, its partitioning is $U_{R}=U/IND(R)$. The number of rules derived from the reduction $R$ is then given by:
\begin{equation}
    num(R,D)=|POS_{R}(U_{D})|=|U_{R}|.
\end{equation}
\end{property}

Obviously, $num(R,D)$ is the cardinality of the positive region of $U_{D}$ relative to the attribute set $R$, which is also the number of rules derived from the reduction $R$. It can be seen that the number of rules is only related to the cardinality of the partitioning of the reduction, and is independent of the number of attribute values in the reduction $R$.

For a reduction $R$, since the final number of rules $num(R,D)$ is equal to its positive region under the decision attribute set $D$, it is also equal to the cardinality of its partitioning $U_{R}$. Therefore, when designing a heuristic algorithm for optimal spatial reduction, we can compare the partitioning $U_{P}=U/IND(P)$ of the current conditional attribute set $P$ with the partitioning $U_{D}$ of the decision attribute set.

When $P$ is not a reduction, Property 1 tells us that adding a conditional attribute $X$ will result in $|U_{P}|\leq |U_{P\cup \{X\}}|$, and the cardinality of the final reduction will also increase, leading to an increase in the number of derived rules. However, selecting conditional attributes solely based on the size of the partitioning's cardinality may lead to the inclusion of many redundant attributes with poor classification capabilities. To address this, we introduce spatial similarity between partitionings to distinguish and reduce the possibility of redundant attributes being included.

For two conditional attribute sets $P$ and $Q$, when $cos(P,D)>cos(Q,D)$, we consider $P$ to be more spatially similar to the partitioning $D$ of the decision attribute set than $Q$, and therefore more likely to yield fewer and more general rules. Moreover, when $cos(P,D)=1$, it indicates that the spatial partitioning of the conditional attribute set $P$ is completely consistent with the partitioning of the decision attribute set $D$. Since both are partitionings of the same universe, this suggests that it is the smallest possible partitioning.

However, spatial similarity alone only reflects the similarity of the spatial structure between two partitionings and does not reflect the actual positive region of the conditional attribute set relative to the decision attribute set. To address this, we introduce attribute importance as an auxiliary criterion to obtain the spatially optimal measure $SigSPS$. Since the operational environment is a compatible decision information system $DIS=\langle U, AT=C\cup D, V,f\rangle$, the introduction of the spatially optimal measure $SigSPS$ ensures convergence of the algorithm. This demonstrates the feasibility of using spatially optimal measures to find reductions with relatively fewer rules.

\subsection{Spatial optimal rough set attribute reduction algorithm}
Algorithm 4 demonstrates the general process of the spatially optimal rough set attribute reduction algorithm. Similar to traditional heuristic algorithms for attribute reduction, it starts by deriving the core of the attribute set. This step typically uses the discernibility matrix approach. When the positive region based on the attribute core is consistent with the positive region based on the conditional attribute set, the core is directly derived as the result of attribute reduction. Otherwise, it calculates the attribute with the highest spatially optimal attribute importance $SigSPS(a,R,D)$ in the conditional attribute set. If the spatially optimal attribute importance is greater than 0, it is added to the core. Otherwise, the attribute with the highest attribute importance in the conditional attribute set is calculated and added to the core. This process continues until the positive region based on the reduced attribute set is consistent with the positive region based on the conditional attribute set, and the reduced attribute set is derived.

\begin{algorithm}[htb]
\renewcommand{\algorithmicrequire}{\textbf{Input:}}
\renewcommand{\algorithmicensure}{\textbf{Output:}}
\caption{Spatial optimal rough set attribute reduction algorithm}
\label{alg:Framwork}
\begin{algorithmic}[1] 
\REQUIRE ~~\\ 
   decision system $DIS=\langle U, AT=C \cup D, V,f\rangle$
\ENSURE ~~\\ 
   attribute set $R$
    \STATE $R= gets minReduction(D)$
    \STATE $While \hspace{0.1cm}POS_{R}(D)\neq POS_{C}(D):$
    \STATE $\hspace{0.5cm} \alpha \gets max SigSPS(a,R,D) (a \subseteq C)$
    \STATE $\hspace{0.5cm} if \hspace{0.1cm} \alpha>0 \hspace{0.1cm} then$
    \STATE $\hspace{1.0 cm} R \gets R \cup \{a\}$
    \STATE $\hspace{0.5cm} else$
    \STATE $\hspace{1.0 cm} a \gets max Sig(a,R,D) (a \subseteq C)$
    \STATE $\hspace{1.0 cm} R \gets R \cup \{a\}$
    \STATE $\hspace{0.5cm} end$
    \STATE $ end$
\STATE \textbf{return} $R$
\end{algorithmic}
\end{algorithm}

Consistent with the traditional rough set attribute reduction algorithm, the time complexity of the spatially optimal rough set attribute reduction algorithm is also divided into two parts: the kernel seeking part and the reduction seeking part. Similarly to the traditional method, the time complexity of the kernel seeking part is $O(|A|\cdot|U|^2)$. The attribute reduction part is consistent with the reduction method based on attribute importance, which is $O(|A|\cdot|U|^2)$.
\section{Experiments}
In order to better illustrate the advantages of the spatial optimal rough set attribute reduction algorithm in obtaining a better spatial similarity and fewer final rules, a comparison experiment with the classic attribute reduction algorithm was designed. The experimental environment is Intel(R) Core(TM) i7-10700F CPU@2.90HZ and GeForce RTX 2060 GPU.

In the experiment, 10 datasets obtained from UCI were used to conduct reduction tests and compare the effectiveness. The MIBARK algorithm and the attribute-oriented rough set approach were selected to compare with the Spatial optimization-based attribute reduction algorithm of rough set (SRS). The number of reduced attributes and the spatial similarity obtained in the end were compared to demonstrate the performance and effectiveness of the algorithm.
\begin{table*}[!hbt]
\centering
\caption{Comparative experimental results\label{tab:table1}}
\renewcommand{\arraystretch}{2.0}
\begin{tabular}{|c|c|c|c|c|c|c|}
\hline
\multirow{2}{*}{DataSet}& \multicolumn{2}{c|}{SRS}&\multicolumn{2}{c|}{MIBARK}&\multicolumn{2}{c|}{attribute-oriented}\\
\cline{2-7}
&Number of attributes&Spatial similarity&Number of attributes&Spatial similarity&Number of attributes&Spatial similarity\\
\hline
Zoo&11&\textbf{0.95}&2&0.20&\textbf{1}&0.28\\
\hline
yellow-small(YS)&\textbf{2}&\textbf{0.84}&4&0.52&\textbf{2}&\textbf{0.84}\\\hline
CarEvaluation(Car)&6&0.03&6&0.03&6&0.03\\\hline
Breast Cancer(BC)&9&0.20&9&0.20&9&0.20\\\hline
Soybean (Small)(SS)&6&\textbf{0.92}&6&0.28&\textbf{2}&0.88\\\hline
adult+stretch(AS)&\textbf{2}&\textbf{0.84}&4&0.52&\textbf{2}&\textbf{0.84}\\\hline
SPECT HEART(SH)&20&0.76&\textbf{15}&0.76&22&0.76\\\hline
lymphography(LY)&13&\textbf{0.40}&10&0.12&\textbf{6}&0.31\\\hline
NPHA-doctor-visits(NDV)&13&0.27&13&0.27&14&0.27\\\hline
primary-tumor(PT)&\textbf{15}&0.48&\textbf{15}&0.48&16&0.48\\
\hline
\end{tabular}
\end{table*}

Table 1 demonstrates the experimental results of the comparative experiments. It can be found that compared with the MIBARK algorithm and the attribute-oriented rough set approach, the Spatial optimization-based attribute reduction algorithm of rough set (SRS) can guarantee a better spatial similarity in most cases. Moreover, on the Zoo dataset, Soybean(small) dataset, and lymphography dataset, it significantly outperforms the traditional methods.

\begin{figure}[h]
    \centering
    \includegraphics[width=3.5in]{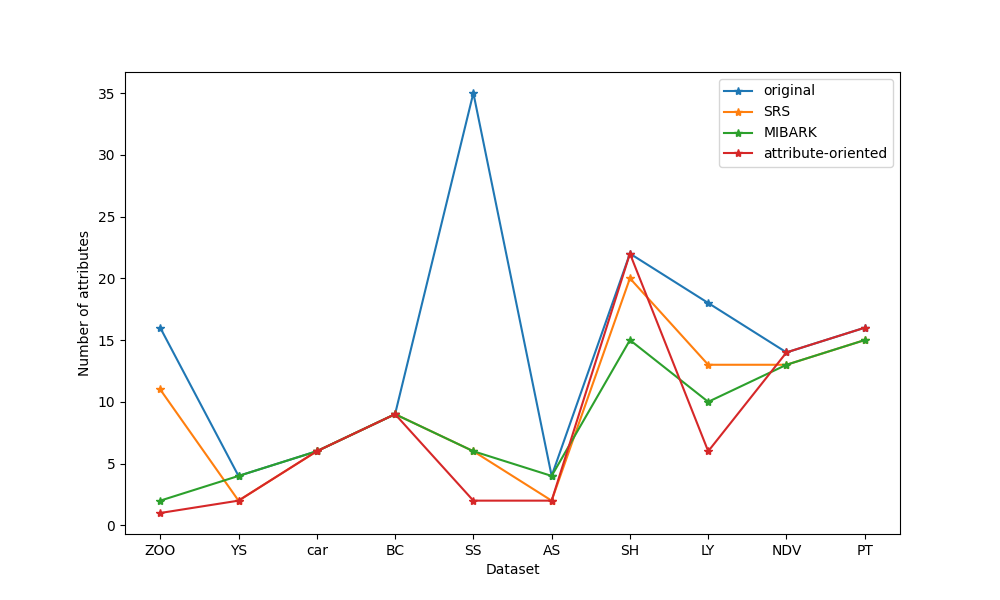}
    \caption{Comparison of Attribute Numbers in Reduced Datasets}
    \label{fig:enter-label}
\end{figure}

Figure 1 compares the number of attributes in the reduced set obtained by the Spatial optimization-based attribute reduction algorithm of rough set (SRS), the MIBARK algorithm, and the attribute-oriented rough set approach with the original number of attributes. It can be observed that apart from the Zoo dataset where there is a significant improvement of 300\% in spatial similarity, the Spatial optimization-based attribute reduction algorithm of rough set (SRS) does not differ much from the MIBARK algorithm and the attribute importance-based heuristic algorithm in terms of the number of attributes. Additionally, on the NDV and PT datasets, the attribute importance-based heuristic algorithm fails to achieve attribute reduction, but both the Spatial optimization-based attribute reduction algorithm of rough set (SRS) and the MIBARK algorithm can achieve a certain level of attribute reduction.

\begin{figure}[h]
    \centering
    \includegraphics[width=3.5in]{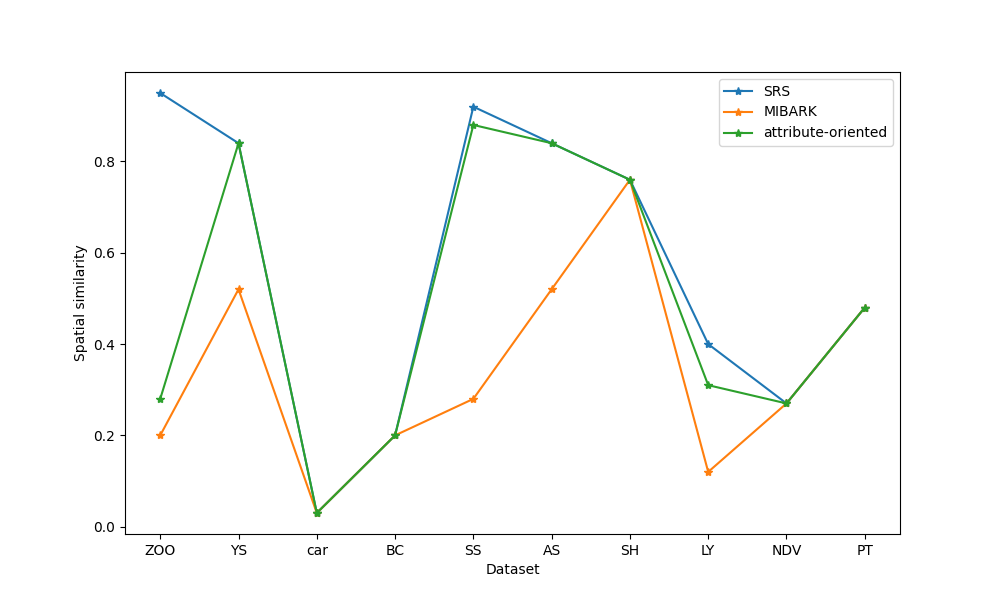}
    \caption{Comparison of Spatial Similarity of Reduced Datasets}
    \label{fig:enter-label}
\end{figure}

Figure 2 demonstrates the spatial similarity comparison among the reduced sets obtained by the Spatial optimization-based attribute reduction algorithm of rough set (SRS), the MIBARK algorithm, and  attribute-oriented rough set approach. It can be found that except for the datasets with consistent reduced sets, when the reduction is inconsistent, the Spatial optimization-based attribute reduction algorithm of rough set (SRS) can achieve better spatial similarity.

\section{Conclusion}
This paper first studies three traditional algorithms for attribute reduction in rough sets, analyzes their time complexity, and finds that they tend to favor attributes with more attribute values when selecting attributes. Subsequently, a spatial optimal-based attribute reduction algorithm for rough sets is proposed to address this issue with traditional methods favoring attributes with more attribute values. The Spatial optimization-based attribute reduction algorithm of rough set (SRS), the goal of obtaining broader and fewer rules is achieved. Finally, the effectiveness of the spatial optimal-based attribute reduction algorithm for rough sets is demonstrated through comparative experiments.\\
\indent {\bf{Data Availability}}: \url{https://archive.ics.uci.edu}\\
\indent {\bf{Code Availability}}: \url{https://github.com/nkjmnkjm/SRS}

\bibliographystyle{ieeetr}
\bibliography{reference}

\begin{thebibliography}{10}

\bibitem{Zalewski_1996}
Pawlak.Z and Janusz, ``Rough sets: Theoretical aspects of reasoning about data,'' {\em Control Engineering Practice}, p.~741–742, May 1996.

\bibitem{Kryszkiewicz1998}
M.~Kryszkiewicz, ``Rough set approach to incomplete information systems,'' {\em Information sciences}, vol.~112, no.~1-4, pp.~39--49, 1998.

\bibitem{Kryszkiewicz1999}
M.~Kryszkiewicz, ``Rules in incomplete information systems,'' {\em Information sciences}, vol.~113, no.~3-4, pp.~271--292, 1999.

\bibitem{CHAUDHURI2021}
A.~Chaudhuri, D.~Samanta, and M.~Sarma, ``Two-stage approach to feature set optimization for unsupervised dataset with heterogeneous attributes,'' {\em Expert Systems with Applications}, vol.~172, p.~114563, 2021.

\bibitem{HU2010}
Q.~Hu, L.~Zhang, D.~Chen, W.~Pedrycz, and D.~Yu, ``Gaussian kernel based fuzzy rough sets: model, uncertainty measures and applications,'' {\em International Journal of Approximate Reasoning}, vol.~51, no.~4, pp.~453--471, 2010.

\bibitem{Greco2001}
S.~Greco, B.~Matarazzo, and R.~Slowinski, ``Rough sets theory for multicriteria decision analysis,'' {\em European journal of operational research}, vol.~129, no.~1, pp.~1--47, 2001.

\bibitem{pawlak1988rough}
Z.~Pawlak, S.~K.~M. Wong, W.~Ziarko, {\em et~al.}, ``Rough sets: probabilistic versus deterministic approach,'' {\em International Journal of Man-Machine Studies}, vol.~29, no.~1, pp.~81--95, 1988.

\bibitem{ziarko1993variable}
W.~Ziarko, ``Variable precision rough set model,'' {\em Journal of computer and system sciences}, vol.~46, no.~1, pp.~39--59, 1993.

\bibitem{yao1992decision}
Y.~Yao and S.~K.~M. Wong, ``A decision theoretic framework for approximating concepts,'' {\em International journal of man-machine studies}, vol.~37, no.~6, pp.~793--809, 1992.

\bibitem{qian2010mgrs}
Y.~Qian, J.~Liang, Y.~Yao, and C.~Dang, ``Mgrs: A multi-granulation rough set,'' {\em Information sciences}, vol.~180, no.~6, pp.~949--970, 2010.

\bibitem{stefanowski1999extension}
J.~Stefanowski and A.~Tsouki{\`a}s, ``On the extension of rough sets under incomplete information,'' in {\em New Directions in Rough Sets, Data Mining, and Granular-Soft Computing: 7th International Workshop, RSFDGrC’99, Yamaguchi, Japan, November 9-11, 1999. Proceedings 7}, pp.~73--81, Springer, 1999.

\bibitem{ye2002new}
D.-Y. Ye and Z.-J. Chen, ``A new discernibility matrix and the computation of a core,'' {\em Acta electronica sinica}, vol.~30, no.~7, pp.~1086--1088, 2002.

\bibitem{hu2008efficient}
Q.~Hu, H.~Zhao, and D.~Yu, ``Efficient symbolic and numerical attribute reduction with neighborhood rough sets,'' {\em Pattern Recognition and Artificial Intelligence}, vol.~21, no.~6, pp.~732--738, 2008.

\bibitem{shu2015incremental}
W.~Shu and W.~Qian, ``An incremental approach to attribute reduction from dynamic incomplete decision systems in rough set theory,'' {\em Data \& Knowledge Engineering}, vol.~100, pp.~116--132, 2015.

\bibitem{greco2001rough}
S.~Greco, B.~Matarazzo, and R.~Slowinski, ``Rough sets theory for multicriteria decision analysis,'' {\em European journal of operational research}, vol.~129, no.~1, pp.~1--47, 2001.

\bibitem{SKOWRON1992}
A.~Skowron and C.~Rauszer, {\em The discernibility matrices and functions in information systems}.
\newblock Intelligent Decision Support. Dordrecht:Springer Netherlands, 1992.

\bibitem{Hu1995}
X.~Hu and N.~Cercone, ``Learning in relational databases: a rough set approach,'' {\em Computational intelligence}, vol.~11, no.~2, pp.~323--338, 1995.

\bibitem{Miao1997}
M.~Duoqian and W.~Jue, ``Information-based algorithm for reduction of knowledge,'' in {\em 1997 IEEE International Conference on Intelligent Processing Systems (Cat. No. 97TH8335)}, vol.~2, (USA: IEEE Publisher), pp.~1155--1158, IEEE, 1997.

\bibitem{JENSEN2005}
R.~Jensen and Q.~Shen, ``Fuzzy-rough data reduction with ant colony optimization,'' {\em Fuzzy sets and systems}, vol.~149, no.~1, pp.~5--20, 2005.

\bibitem{wang2019feature}
C.~Wang, Y.~Huang, M.~Shao, Q.~Hu, and D.~Chen, ``Feature selection based on neighborhood self-information,'' {\em IEEE Transactions on Cybernetics}, vol.~50, no.~9, pp.~4031--4042, 2019.

\bibitem{wang2021attribute}
C.~Wang, Y.~Huang, W.~Ding, and Z.~Cao, ``Attribute reduction with fuzzy rough self-information measures,'' {\em Information Sciences}, vol.~549, pp.~68--86, 2021.

\bibitem{wang2019fuzzy}
C.~Wang, Y.~Wang, M.~Shao, Y.~Qian, and D.~Chen, ``Fuzzy rough attribute reduction for categorical data,'' {\em IEEE Transactions on Fuzzy Systems}, vol.~28, no.~5, pp.~818--830, 2019.

\bibitem{xia2020gbnrs}
S.~Xia, H.~Zhang, W.~Li, G.~Wang, E.~Giem, and Z.~Chen, ``Gbnrs: A novel rough set algorithm for fast adaptive attribute reduction in classification,'' {\em IEEE Transactions on Knowledge and Data Engineering}, vol.~34, no.~3, pp.~1231--1242, 2020.

\bibitem{sang2021incremental}
B.~Sang, H.~Chen, L.~Yang, T.~Li, and W.~Xu, ``Incremental feature selection using a conditional entropy based on fuzzy dominance neighborhood rough sets,'' {\em IEEE Transactions on Fuzzy Systems}, vol.~30, no.~6, pp.~1683--1697, 2021.

\bibitem{ewees2020improved}
A.~A. Ewees, M.~Abd~Elaziz, R.~M. Arafa, and R.~M. Ghoniem, ``Improved approach based on fuzzy rough set and sine-cosine algorithm: A case study on prediction of osteoporosis,'' {\em IEEE Access}, vol.~8, pp.~203190--203202, 2020.

\end{thebibliography}

\vfill

\end{document}